\documentclass[10pt,journal,compsoc]{IEEEtran}
\usepackage{color}
\usepackage{graphicx}
\usepackage{url}
\usepackage{subfigure}

\ifCLASSOPTIONcompsoc
  \usepackage[nocompress]{cite}
\else
  \usepackage{cite}
\fi
\ifCLASSINFOpdf
\else
\fi
\begin{document}
\title{DAC-SDC Low Power Object Detection Challenge for UAV Applications}

\author{Xiaowei Xu,~\IEEEmembership{Member,~IEEE,}
        Xinyi Zhang,~\IEEEmembership{Student Member,~IEEE,}
        Bei Yu,~\IEEEmembership{Senior Member,~IEEE,}
        Xiaobo Sharon Hu,~\IEEEmembership{Fellow,~IEEE,}
        Christopher Rowen,~\IEEEmembership{Fellow,~IEEE,}
        Jingtong Hu,~\IEEEmembership{Member,~IEEE,}
        and Yiyu Shi,~\IEEEmembership{Senior Member,~IEEE}% <-this % stops a space
\IEEEcompsocitemizethanks
{\IEEEcompsocthanksitem Xiaowei Xu, Xiaobo Sharon Hu and Yiyu Shi are with the Department of Computer Science and Engineering, University of Notre Dame, IN, 46556.\protect\\
E-mail: xxu8@nd.edu, shu@nd.edu, yshi4@nd.edu
\IEEEcompsocthanksitem Bei Yu is with the Department of Computer Science and Engineering, The Chinese University of Hong Kong, Hong Kong.\protect\\
E-mail: byu@cse.cuhk.edu.hk
\IEEEcompsocthanksitem Xinyi Zhang and Jingtong Hu are with the Department of Electrical and Computer Engineering,
University of Pittsburgh, PA, 15261.\protect\\
E-mail: xinyizhang@pitt.edu, jthu@pitt.edu
\IEEEcompsocthanksitem Christopher Rowen is the CEO of Cognite Venture,
210 Mission Street, Santa Cruz, CA 95060.\protect\\
E-mail: rowen@cogniteventures.com
}% <-this % stops an unwanted space

\thanks{Manuscript received July 31, 2018; revised August 26, 2018.}}


\markboth{Journal of \LaTeX\ Class Files,~Vol.~14, No.~8, August~2018}%
{Shell \MakeLowercase{\textit{et al.}}: Bare Demo of IEEEtran.cls for Computer Society Journals}



\IEEEtitleabstractindextext{%
\begin{abstract}
%The Design Automation Conference system design contest (DAC-SDC) lower power object detection challenge is a benchmark in object detection on 95 categories and 150k images under the view of unmanned aerial vehicles (UAV) perspective.
%The challenge has been run on the 55th Design Automation Conference in 2018, attracting more than 110 entries from all over the world.
%This paper describes the dataset and evaluation procedure.
%The methods and optimization of the entries are discussed under this particular UAV applications, and lessons are provided with detailed analysis.
%The paper concludes with directions for future improvements.
The 55th Design Automation Conference (DAC) held its first System Design Contest (SDC) in 2018.
SDC'18 features a lower power object detection challenge (LPODC) on designing and implementing novel algorithms based object detection in images taken from unmanned aerial vehicles (UAV).
The dataset includes 95 categories and 150k images, and the hardware platforms include Nvidia's TX2 and Xilinx's PYNQ Z1.
DAC-SDC'18 attracted more than 110 entries from 12 countries.
This paper presents in detail the dataset and evaluation procedure.
It further discusses the methods developed by some of the entries as well as representative results.
The paper concludes with directions for future improvements.

\end{abstract}

}

\maketitle

\IEEEdisplaynontitleabstractindextext

% \fi
\IEEEpeerreviewmaketitle

\IEEEraisesectionheading{\section{Introduction}\label{sec:introduction}}


\IEEEPARstart{T}
%{he} Design Automation Conference system design contest (DAC-SDC) lower power object detection challenge (LPODC) consists of two primary segments: a publicly available dataset of images and annotation with evaluation software; and a competition and session hosted on the 55th Design Automation Conference (DAC).
%The challenge provides a unified platform to develop and compare state-of-the-art object detection algorithms, and discusses the lessons learned from these participated entries.
The 55th Design Automation Conference (DAC) held its first System Design Contest (SDC) in 2018 which features a lower power object detection challenge (LPODC) on designing and implementing novel algorithms based object detection in images taken from unmanned aerial vehicles (UAV).
This challenge provides a unified platform to develop and compare state-of-the-art object detection algorithms, and discusses the lessons learned from these participated entries.

The LPODC at DAC-SDC'18 focuses on unmanned aerial vehicles (UAV) applications as such applications have stringent accuracy, real-time, and energy requirements.
Specifically, first, the LPODC task is to detect a single object of interest, one of the most important tasks in UAV applications \cite{torres2015automatic}.
%Second, unlike general computer visual challenges such as ImageNet \cite{russakovsky2015imagenet} and  PASCALVOC  Dataset \cite{everingham2010pascal}, which focus only on accuracy, LPODC evaluates the final performance based on a combination of throughput, power, and detection accuracy which takes into full consideration the features of UAV applications: real-time processing, energy-constraint embedded platform, and detection accuracy.
Second, different from general computer visual challenges, such as ImageNet \cite{russakovsky2015imagenet} and PASCAL VOC dataset \cite{everingham2010pascal}, which focused
only on accuracy, LPODC evaluates the final performance based on a combination of throughput, power, and detection accuracy. Thus, LPODC takes into full consideration the features of UAV applications: real-time processing, energy-constrained embedded platform, and detection accuracy.
%general computer visual challenges such as ImageNet \cite{russakovsky2015imagenet} and PASCALVOC dataset \cite{everingham2010pascal} focused only on accuracy.
%LPODC evaluates the final performance based on a combination of throughput, power, and detection accuracy which takes into full consideration the features of UAV applications: real-time processing, energy-constraint embedded platform, and detection accuracy.
Third, the images of the dataset are all captured from actual UAVs which reflect the real circumstances and problems of UAV applications.
Fourth, LPODC provides two hardware platforms: embedded GPU (Jetson TX2 from Nvidia \cite{nvidia}) and FPGA SoC (PYNQ Z-1 board from Xilinx \cite{xilinx}) to all the participating teams to choose from for their implementations.
Note that GPUs and FPGAs are widely adopted for energy-efficient processing on UAVs \cite{chao2013survey}.

The publically released dataset contains a large quantity of manually annotated training images, while the testing dataset is withheld for the evaluation purpose.
There are a total of 150k images provided by a UAV company DJI \cite{dji}.
Participating teams trained their models/algorithms with the training dataset, and sent the trained models/algorithms to the organizers to get the final testing results including throughput, energy, and detection accuracy.
Such evaluation was performed at the end of each month and the detailed rank was released then.
The final rank was released at the end of the competition and the top-3 entries from both GPU and FPGA categories were invited to present their work at a technical session at DAC.

This paper describes the LPODC in detail including: the task, the evaluation method and the dataset.
Furthermore, a comprehensive discussion of the methods and results of the top-3 entries from both GPU and FPGA categories is presented to provide insights and  rich lessons for future development of object detection algorithms especially for UAV applications.
Particularly, we will elaborate hardware-software co-design for efficient processing on embedded platforms.

The training dataset, the source codes of the top-3 entries of both GPU and FPGA categories, and additional information about this challenge can be found at \url{www.github.com/xyzxinyizhang/2018-DAC-System-Design-Contest}.

\subsection{Related Work}
In this section we briefly discuss the related work about benchmark image datasets for object detection.
As segmentation datasets can also be used for object detection, some widely-used segmentation datasets are also included.

Most of the datasets for object detection contain common photographs.
LabelMe \cite{russell2008labelme} has 187k images each of which contains multiple objects annotated with bounding polygon.
It also provides a web-based online annotation tool for easy contribution by the public.
Like LabelMe, PASCAL VOC dataset \cite{everingham2010pascal} further increases the number of images to 500k.
ImageNet \cite{russakovsky2015imagenet} is one of the most popular datasets in computer vision community, and contains more than 14 million images.
It was primarily for classification in 2010 and extended to support object detection and scene recognition in 2013.
Compared with ImageNet, SUN database \cite{xiao2010sun} mainly focuses on scene recognition and contains about 131k images.
Microsoft Common Objects in Context (COCO) \cite{lin2014microsoft} contains complex everyday scenes of common objects in their natural context and has 2.5 million images.
Open Images \cite{openimages} contains more than 9 million real-life images within 6,000 categories which is much larger than that of ImageNet (about 1,000).

There are also some datasets for specific applications, and the images are taken from particular views.
KITTI vision benchmark dataset \cite{Geiger2012CVPR} is specific for autonomous driving, and the images are taken from a autonomous driving platform in a mid-size city.
FieldSAFE \cite{kragh2017fieldsafe} is for agriculture application and has approximately 2 hours of raw sensor data from a tractor-mounted sensor system in a grass mowing scenario.
DOTA \cite{xia2018dota} focuses on aerial applications, and all the images are captured from cameras on aircrafts.

The dataset in this paper is specific for UAV applications.
The images in the dataset are taken from UAVs which operates on a much lower height than general aircrafts in DOTA.
The associated environment in the images is also much more complex than that in DOTA.

\subsection{Paper Layout}

We have given an overview of the LPODC at DAC'18.
The rest of the paper is organized as follows. In Section 2,
%The organization of the paper is as follows.
%The overview of the challenge is presented in Section 1,
%and
the challenge task and its evaluation method, as well as the provided two hardware platforms are described.
The details of the dataset and its analysis are given in Section 3.
The analysis and discussion of the methods for object detection of GPU and FPGA entries are presented in Section 4.
Section 5 details the results of the GPU and FPGA entries.
We conclude the paper with some discussions of the challenge and possible improvements.

\section{LPODC Task and Evaluation Criteria}
In this section, the details of the challenge task are presented, followed with an introduction of the hardware platforms and the evaluation method.

\subsection{Object Detection}
The LPODC task is to perform single object detection in each image with an axis-aligned bounding box indicating the object's position and scale.
As the challenge is targeted at UAV applications, there are several aspects that need to be emphasized.
%First, the object detection task is to find exactly the same object in the testing images as that in the training dataset rather than objects belong to the same category.
First, the object detection task is to locate a specific object from the training dataset, rather than objects from a training category.
For example, if images containing person A are in the training dataset, then the task is to detect person A rather than other persons.
More details about the detection objects are discussed in Section 3.
Second, the object detection task needs to be executed with high throughput and high accuracy which are required by UAV applications.
This requirement is achieved through the weighting of throughput in the scoring system, discussed in Section 2.3.
%by giving a penalty to low throughput implementations, and the details are discussed in Section 2.3.

\subsection{Hardware Platforms}

%\subsubsection{FPGA}
In this challenge, two hardware platforms, either FPGA or GPU, were provided to the participating teams from the challenge sponsors Xilinx \cite{xilinx} and Nvidia \cite{nvidia}, respectively.
Particularly, the \textbf{FPGA platform} is Xilinx PYNQ Z-1 board which is an embedded systems based platform combining Zynq system and Python~\cite{janssen2017dynamic}.
%PYNQ is an embedded systems targeted platform which combine Zynq system and Python~\cite{janssen2017dynamic}.
Participants are allowed to use Cortex-A9 processor and ZYNQ XC7Z020-1CLG400C on the platform to realize their solutions to the challenge.
The embedded FPGA chip contains 53K 6-input look-up-tables, 220 DSPs, and 630KB fast block RAM.
A 512MB DDR3 memory with 16-bit bus at 1,050Mbps is also embedded on the platform which can be accessed by both the processor and the FPGA.
The power consumption of the FPGA platform is about 1-4 watts.

The \textbf{GPU platform} is Nvidia Jetson TX2, which is an embedded AI computing device.
This GPU is very powerful with a 6-core CPU (Dual-core NVIDIA Denver2 + quad-core ARM Cortex-A57), a 256-core Pascal GPU, and 8GB LPDDR4 DRAM.
It can provide more than 1TFLOPS of FP16 compute performance in less than 7.5 watts of power.
Note that both hardware platforms target low-power embedded computation and are suitable for UAV applications.

\begin{figure*}
  \includegraphics[width=1\textwidth]{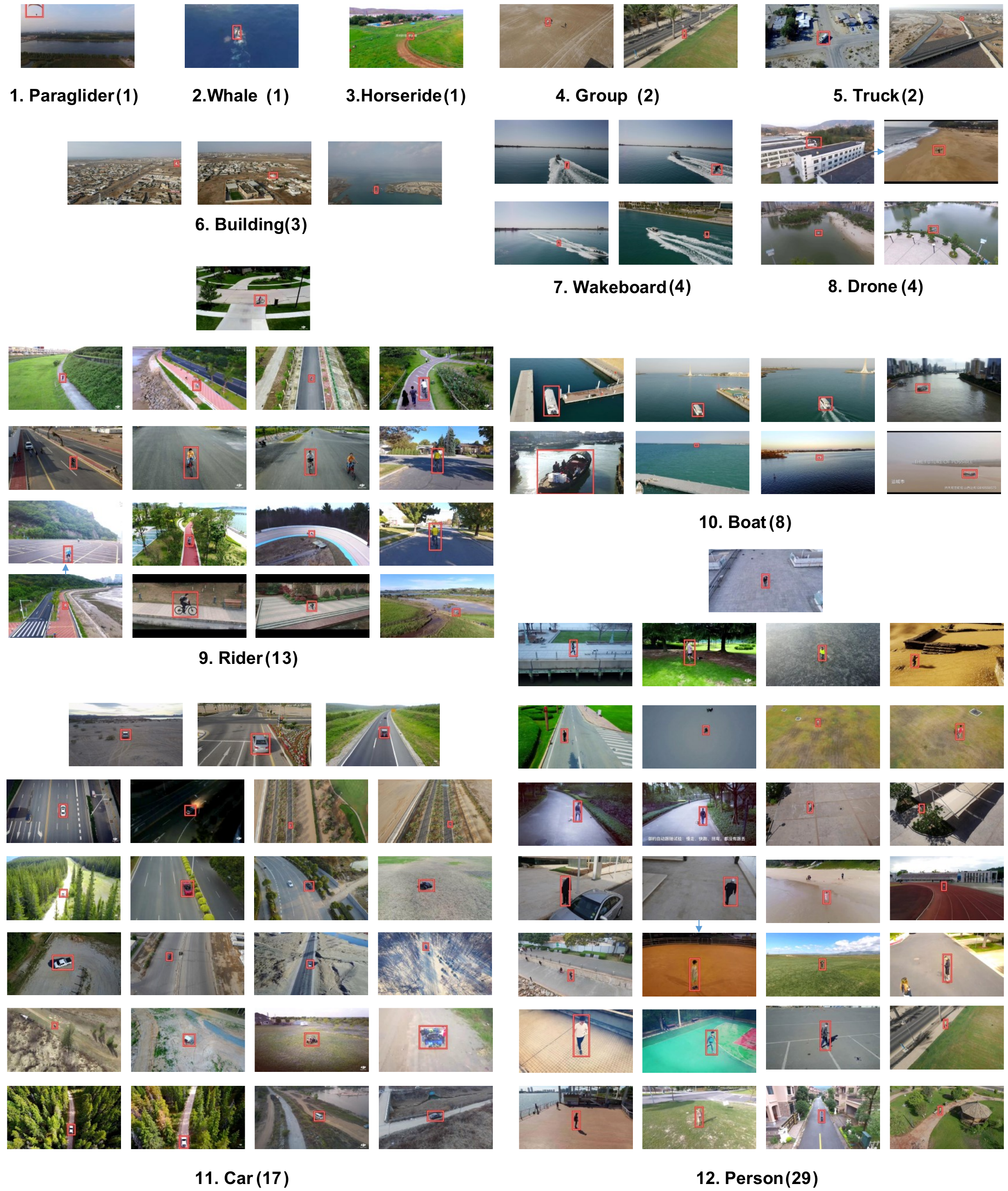}
  \vspace{-11pt}
  \caption{Overview of the dataset provided by DJI. There are 12 categories, each of which includes several sub-categories as indicated in the bracket, and there are totally 95 sub-categories. Note that there is only one object in each image.}
    \vspace{-1pt}
  \label{fig:dataset}
  \vspace{-16pt}
\end{figure*}

\subsection{Evaluation Method}

The evaluation for the challenge is based on detection accuracy, throughput, and energy consumption.
As energy consumption is relatively straightforward, the following content mainly discusses the metrics for accuracy and throughput.

The metric for object detection accuracy is Intersection over Union (IoU).
Suppose there are two bounding boxes (BB): a predicted BB and the ground-truth BB.
Then the accuracy or IoU of the predicted BB is the ratio between the area of the union of the predicted BB and the ground-truth BB and the area of the overlap encompassed by both the predicted BB and the ground-truth BB, i.e.,
\begin{equation}\label{distance}
%d(c_i'',t_j'')=(c_i''-t_j'')^2.
IoU = \frac{BB_{p}\cap BB_{g}}{BB_{p}\cup BB_{g}},
\end{equation}
where $BB_{p}$ and $BB_{g}$ are the areas of the predicted and ground-truth BBs, respectively.
Note that the challenge only cares the IoU results, but does not care the object categories.

The metric for throughput is frames per second (FPS).
For real-time processing in UAV applications, the minimum throughput requirement in this challenge was set to 20 FPS on the GPU platform and 5 FPS on the FPGA platform.
If the FPS is lower than the requirement, then a penalty to IoU is added, i.e.,
\begin{equation}\label{penalty}
%d(c_i'',t_j'')=(c_i''-t_j'')^2.
IoU_{r} = IoU_{m}\times (min(FPS_{m}, FPS_{r}))/FPS_{r},
\end{equation}
where $IoU_{r}$ and $IoU_{m}$ are the actual and measured IoUs, respectively, and $FPS_{r}$ and $FPS_{m}$ are the required and measured FPSs, respectively.
The $min$ function outputs the minimum one of its inputs.

The final score is a combination of accuracy, throughput, and energy consumption.
Suppose there are $I$ registered entries and the dataset contains $K$ evaluation images.
Let $IoU_{i,k}$ be the IoU score of image $k$ ($k \leq K$) for entry $i$ ($i\leq I$).
Then the IoU score $R_{IoU_{i}}$ for entry $i$ is computed as
\begin{equation}\label{eval1}
R_{IoU_{i}} = \frac{\sum_{k=1}^{K}IoU_{i,k}}{K}.
\end{equation}
Let $E_i$ be the energy consumption of processing all $K$ images for entry $i$.
Let $\bar{E_I}$ be the average energy consumption of the $I$ entries.
$\bar{E_I}$ is computed as
\begin{equation}\label{eval2}
\bar{E_I} = \frac{\sum_{i=1}^{I}E_{i}}{I}.
\end{equation}
Then the energy consumption score $ES_i$ for entry $i$ is
\begin{equation}\label{eval3}
{ES_i} = max\{0, ~1+0.2\times log_x\frac{\bar{E_{I}}}{E_i}\},
\end{equation}
where $x$ is set to 2 and 10 for FPGA category and GPU category, respectively.
Through profiling, we find that the energy-performance Pareto frontier of the GPU platform exhibits much smaller gradient than that of the FPGA platform, reflecting the fact that the FPGA is more resource constrained than the GPU and thus its performance is much more sensitive to energy consumption. As such, we stress more on the accuracy for the GPU, while more on the energy for the FPGA.
The final total score $TS_i$ for entry $i$ is
\begin{equation}\label{eval4}
{TS_i} = R_{IoU_i}\times (1+{ES_i}).
\end{equation}

The factor 0.2 in Equation (\ref{eval3}) is set based on the estimated range of energy consumption variation in participating teams.
Through our profiling of the platform, we anticipate that a team will normally have an energy consumption from 1/4x to 4x the average of all teams for both GPU and FPGA categories.
%we anticipate that $log_{x}\bar{E_I}/E_{i}$ will take a range of -0.25 to 2, and
As such, $log_{x}\bar{E_I}/E_{i}$ will take a range of -0.25 to 2, and a factor of 0.2 will make $ES_{i}$ in the range of 0.6 to 1.4.
Thus, the final (1+$ES_{i}$) in Equation (\ref{eval4}) is in the range of 1.6 to 2.4 which is appropriate to act as a reward factor for energy efficiency.
The $max$ function and the addition of 1 to $ES_{i}$ are for teams with extremely low performance on energy efficiency.
In such condition, $log_{x}\bar{E_I}/E_{i}$ is a large negative number, and the $max$ function ensures that $ES_{i}$ is not a negative number but zero.
Then the addition of 1 to $ES_{i}=0$ further ensures that $TS_{i}$ can still be graded based on accuracy with no multiplying rewarding factor for energy efficiency rather than being frozen to zero even if very high IoUs are obtained.

%As powerful GPUs are usually for complex tasks, while resource-constraint FPGAs are usually for only one specific application, higher weights ($x$=10 and 2) are given to accuracy and energy for GPUs and FPGAs, respectively.

%This is to capture the characteristic of UAV applications where accuracy, throughput and energy/power are all very important.
%It can be noted from the equations that when the real-time processing requirement is met, the detection accuracy (IoUs) plays a more important role in the final evaluation.

\section{Dataset}\label{dataset}

{\color{black}{As shown in Fig. \ref{fig:dataset}, the adopted dataset from DJI \cite{dji} contains 12 categories of images and 95 sub-categories.
For each sub-category, 70\% of the images are provided for training and 30\% are reserved for evaluation.}}
%It should be highlighted that there is a significant difference between our dataset and other general purpose dataset such as
%In each class, the object is captured in a UAV view and with different point of view. The size of the object varies from image to image in a class.
It should be highlighted that compared with existing general purpose datasets such as ImageNet \cite{russakovsky2015imagenet} and  PASCAL VOC dataset \cite{everingham2010pascal},  the object is captured in a UAV view and with different points of view.

\begin{figure}
\centering
  \includegraphics[width=1\columnwidth]{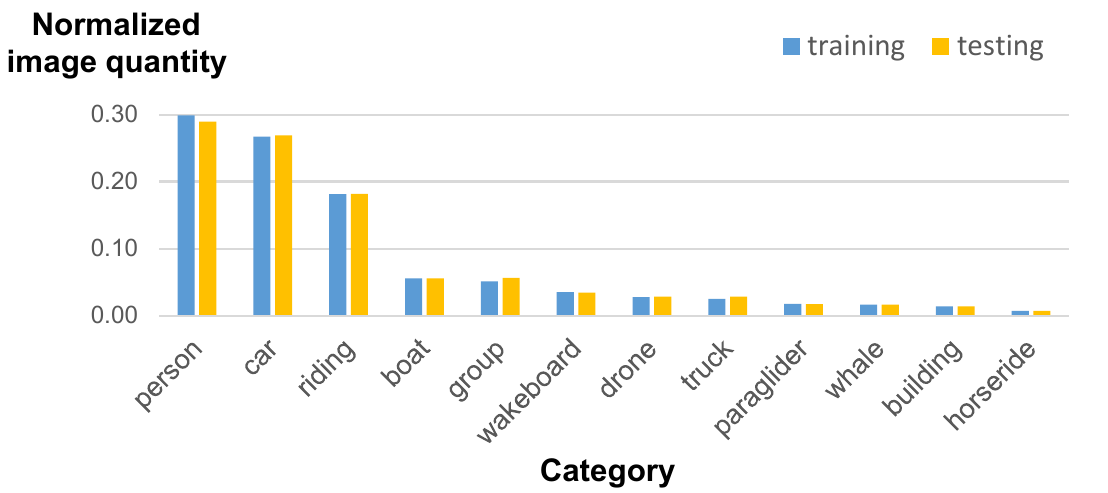}
  \vspace{-25pt}
  \caption{Distributions of the training and testing datasets with respect to image categories.}
  \vspace{-5pt}
  \label{fig:dataset_category}
\end{figure}

\begin{figure}
\centering
  \includegraphics[width=1\columnwidth]{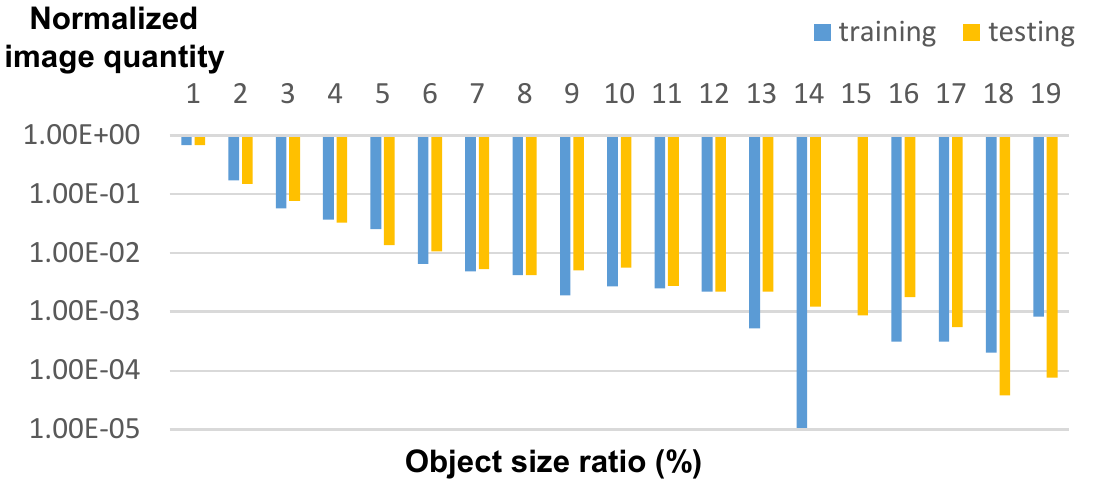}
  \vspace{-25pt}
  \caption{Distributions of the training and testing datasets with respect to object size ratio.}
  \vspace{-5pt}
  \label{fig:dataset_size}
\end{figure}

\begin{figure}
\centering
  \includegraphics[width=1\columnwidth]{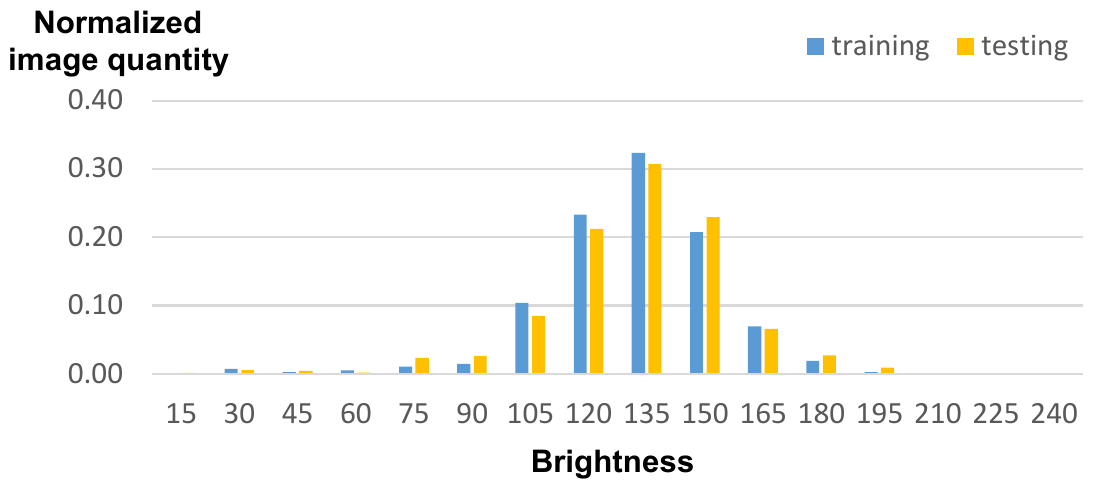}
  \vspace{-25pt}
  \caption{Distributions of the training and testing datasets with respect to image brightness.}
  \vspace{-5pt}
  \label{fig:dataset_brightness}
\end{figure}

\begin{figure}
\centering
  \includegraphics[width=1\columnwidth]{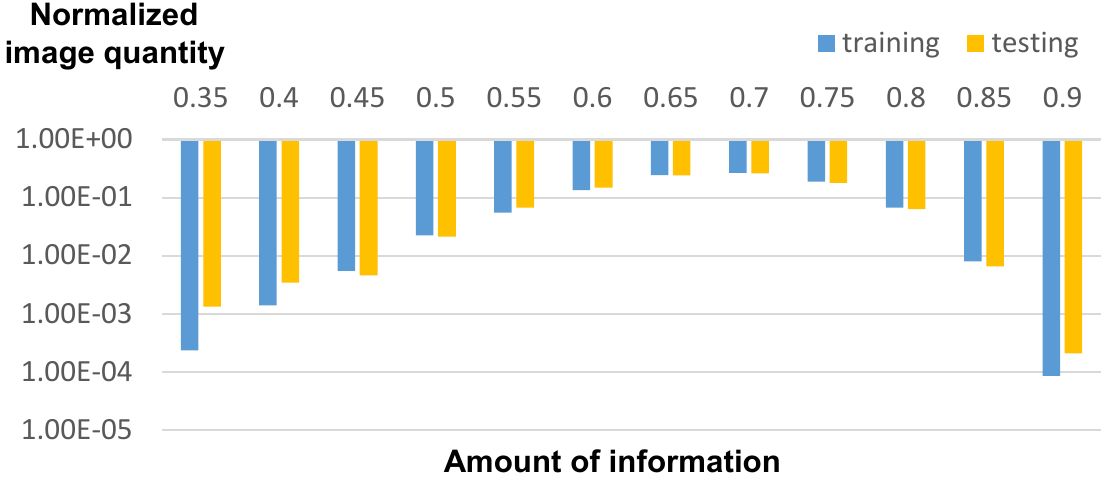}
  \vspace{-25pt}
  \caption{Distributions of the training and testing datasets with respect to amount of information.}
  \label{fig:dataset_texture}
\end{figure}

The distributions of the training and testing datasets with respect to category, object size ratio, image brightness and amount of information are shown in Figs. \ref{fig:dataset_category}-\ref{fig:dataset_texture}.
Here, object size ratio is the ratio of the object size to the image size.
The brightness of a pixel is defined as in Equation (\ref{brightness})\cite{sanmorino2012clustering} (r, g, b are the three channels of images), where the image brightness is the average brightness of all its pixels.
The amount of information is defined as the average pixel entropy of the object where the pixel entropy is calculated in a 5$\times$5 region.
%Note that for clear demonstration,

\begin{equation}\label{brightness}
%d(c_i'',t_j'')=(c_i''-t_j'')^2.
brightness = \sqrt{0.241\times r^2+0.691\times g^2+0.068\times b^2}
\end{equation}

%of the object varies from image to image in a class
Fig. 2 shows that the ratio of quantity of training and testing images in different categories are almost the same, which is manually segmented to achieve a good balance of training and testing dataset.
The categories \textit{person}, \textit{car}, and \textit{rider} contain much more images than others as they contain more sub-categories than others.
It can be noticed that there is also a good balance between the training dataset and the testing dataset for different object sizes except for some large object size ratio as shown in Fig. 3.
As the object size ratio increases, the image quantity decreases.
Note that the average object size ratio in ILSVRC is 17\% and in PASCAL VOC is 20\%.
However, in this dataset, most of the images have an object size of 1-2\% of the captured images (640x360), which is the main character of UAV-view images.
This good balance still holds for image brightness and amount of information as shown in Fig. \ref{fig:dataset_brightness} and Fig. \ref{fig:dataset_texture}.
Both the two distributions in the two figures have the same shape: most of the images have a moderate brightness/amount of information, while many fewer images contain too large or too small brightness/amount of information, which are like a Gaussian distribution.

%Among all images, most of the images has a object size around 1.8\% of the captured images (640x360), while other dataset such as xxxxx has much bigger object size, xxxxxxxxxxxxxxxx.
%{\color{red}{!!will show 4 figs about quantity vs. (1)category(2)object size(3)brightness(4)texture, and discuss them.!!}}

%Note: The dataset provided for participants to download contains 70\% of the total dataset provided by our sponsor. The remaining 30\% of the dataset is reserved for our evaluation. We will ONLY use the reserved dataset to evaluate and rank all the entries.

\section{Methods for Object Detection}

In this section, the analysis and discussion of the methods for object detection reported by the representative GPU and FPGA entries are presented.
The details of the top-3 entries of GPU and FPGA categories are discussed, and statistical significance analysis is also presented.

\subsection{GPU}
\begin{figure}
\centering
\subfigure{
\label{fig:gpu_summary_1}
\includegraphics[width=0.23\textwidth]{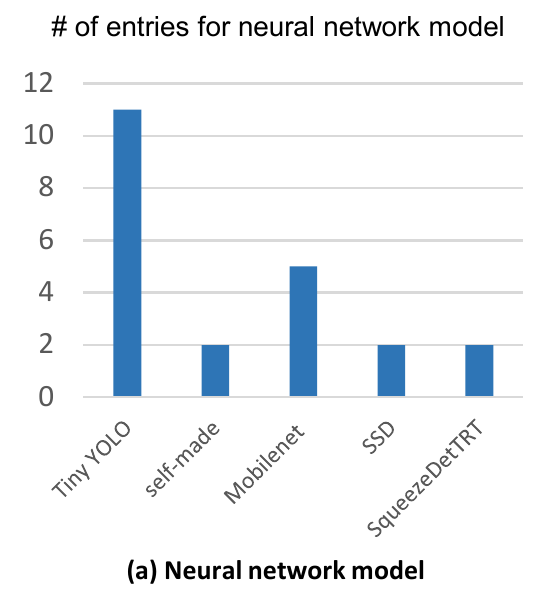}}
\subfigure{
\label{fig:gpu_summary_2}
\includegraphics[width=0.23\textwidth]{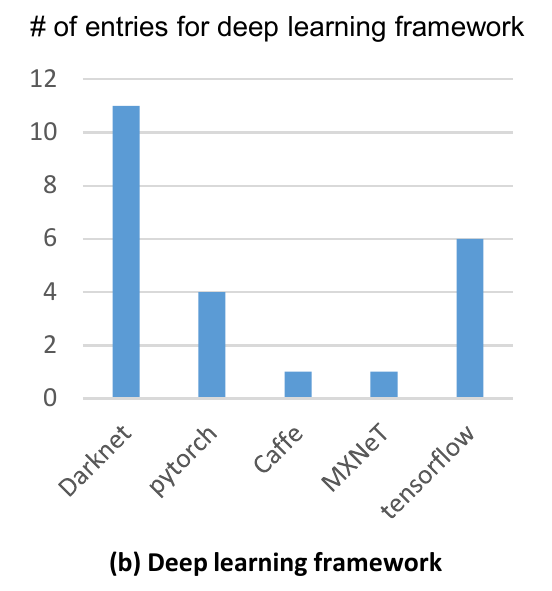}}
\vspace{-5pt}
\caption{Number of entries using (a) neural network models and (b) deep learning frameworks.}
\label{fig:gpu_summary}
\vspace{-15pt}
\end{figure}

A total of 24 out of the 53 participating teams successfully implemented their designs under the GPU category, and all of them adopted deep learning approach.
The distributions of used neural network models and deep learning frameworks are shown in Fig. \ref{fig:gpu_summary}.
Tiny YOLO \cite{redmon2016you} is the most widely used model in the contest, and a majority of the entries achieve high performance by adding some revision to the network structure.
Darknet is the most popular deep learning framework as tiny YOLO is originally implemented in Darknet.
The top three entries \textbf{ICT-CAS}, \textbf{DeepZ}, and \textbf{SDU-Legend} all adopted the YOLO model as the base design and improved it with structure and computation optimization.
Note that the image size is 640$\times$360 in the challenge, and ICT-CAS and SDU-Legend resized it to improve accuracy and throughput.

\begin{figure}
\centering
  \includegraphics[width=1\columnwidth]{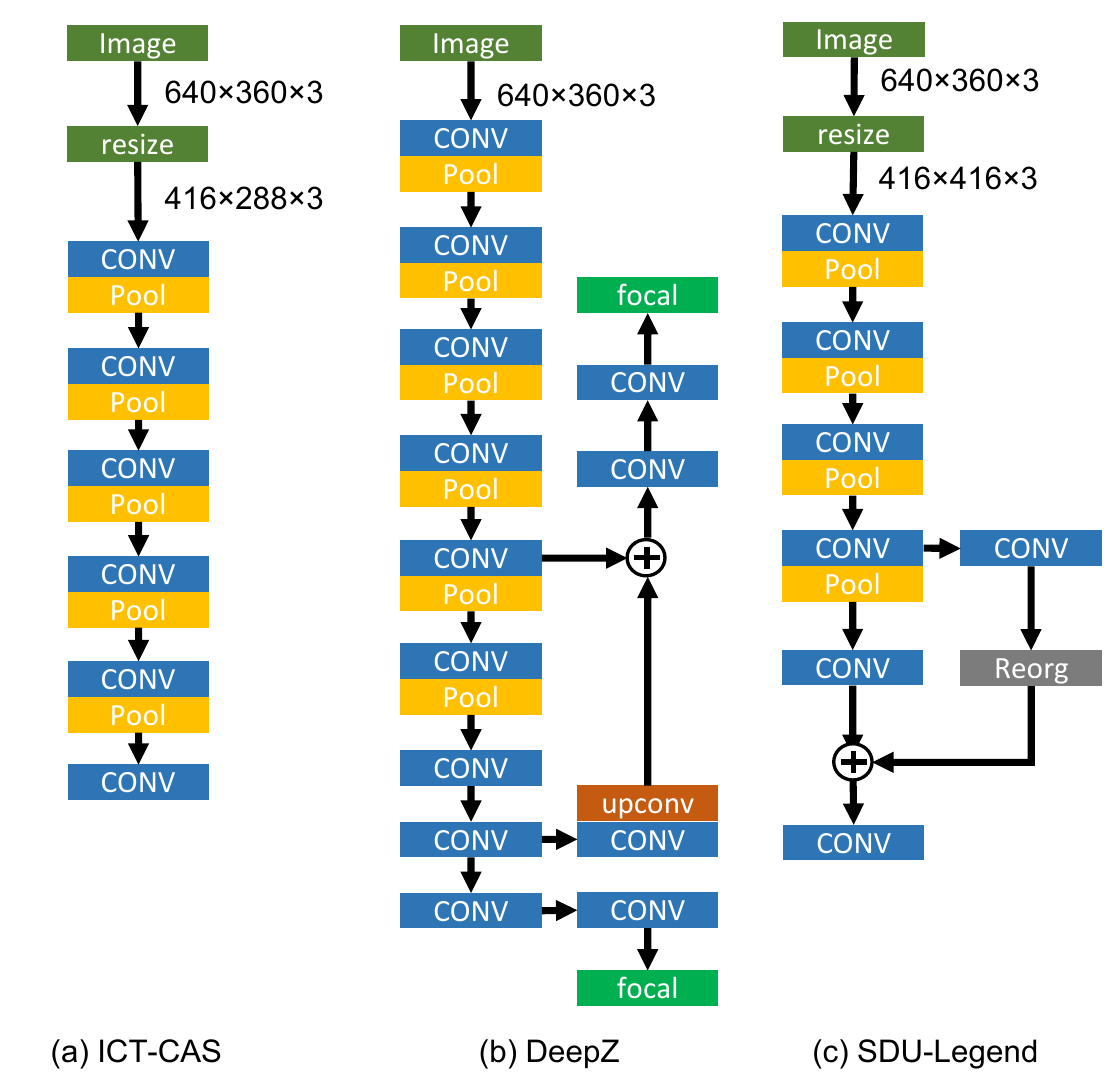}
  \caption{Neural network structure of the top-3 GPU entries: (a) ICT-CAS, (b) DeepZ, and (c) SDU-Legend.}
  \label{fig:Network_GPU}
\end{figure}

\textbf{{ICT-CAS}} adopted the original tiny YOLO as their network structure as shown in Fig. \ref{fig:Network_GPU}(a), and deployed the Tucker decomposition, hard example mining and low-bits computation for fast and accurate processing.
%, while in pursuit of speed, we tried Detection Framework optimization selection, lightweight CNN for feature extractor, tucker decomposition and precision scaling. In Detection Framework selection, we tested Mask/Faster R-CNN, YOLOv2/v3, SSD to find out best Detection Framework for the detection task. As for lightweight CNN, We tried mobilenetv1/v2, shufflenet and Xception as feature extractor to reduce computation and memory consumption.
In Tucker decomposition, they tested different decomposition parameters for optimal precision and throughput.
By extracting the hard examples, re-training was performed to increase accuracy.
In order to speed up the inferencing stage, they deployed half precision float point computation to reduce computation complexity and power consumption.
%In parameter exploration, we mainly tested input size and anchor box parameter.
In the TX2 GPU platform, this entry also adopted TensorRT \cite{tensorrt} as the inference optimizer to speed up the inference.

%Due to the speed limitation of 20FPS, we started with YOLOv2-Tiny as the backbone network for feature extraction and a detection network for candidate bounding box generation. However, with such a simple backbone model, we were soon faced with the challenges of tiny objects, occlusions and distraction from the provided data set. In order to tackle to the aforementioned challenges, we investigated various network architectures for both inference and training.
\textbf{DeepZ} implemented their own network structure as shown in Fig. \ref{fig:Network_GPU}(b).
It combined Feature Pyramid Network \cite{lin2017feature} to fuse fine-grained features with strong semantic features to enhance the ability in detecting small objects. Meanwhile, DeepZ utilized focal loss function to mitigate the imbalance between the single ground truth box and the candidate boxes in the training phase, thereby partially resolving occlusions and distractions.
%With the combined techniques, we achieved the inference network as shown in the figure with an accuracy improvement of ~ 4.2\% while maintaining pretty much the same speed.

\textbf{SDU-Legend} focused on both neural network and architectural level optimization to achieve better balance between system performance and detection accuracy.
SDU-Legend chose YOLO v2 as the starting design point, and performed design space exploration to choose the key training parameters including anchors, coordinates scale in the loss function, batch size, and learning rate policy.
Moreover, SDU-Legend reduced YOLO v2 network architecture from 32 layers to 27 layer (as shown in Fig. \ref{fig:Network_GPU}(c)), and decreased the downsampling rate to strengthen the performance on small targets.
At the architectural level, SDU-Legend aimed at balancing the workload between the GPU and the CPU by executing the last 2 layers on the CPU.
SDU-Legend utilized the half data type (16-bit float) instead of 32-bit float to improve the memory throughput and reduce computation cost with minimum loss in accuracy. %Finally, the object detection system from entry ``SDU-Legend’’ achieves IoU of 0.6847, 10.3w power, and 23.64 FPS.

\subsection{FPGA}

There are a total of seven out of the 61 participating teams that successfully implemented their designs on the FPGA platform provided.
Among these entries, only entry \emph{{TGIIF}} adopted Verilog hardware description language for compact FPGA implementation, while the rest utilized high-level synthesis for fast FPGA implementation.
%synthesis and only entry \emph{\textbf{TGIIF}} adopted Verilog as the programming language. And the majorities of the entries adopted a fixed-point operation in designs to make it hardware friendly.

The top three entries \emph{\textbf{TGIIF}}, \textit{\textbf{SystemsETHZ}}, and \textbf{\textit{iSmart2}} adopted Convolution Neural Network and used variations on the models such as Single Shot MultiBox Detector (SSD)~\cite{liu2016ssd}, SqueezeNet~\cite{iandola2016squeezenet}, MobileNet~\cite{howard2017mobilenets}, and Yolo~\cite{redmon2016you}.
From the original models, the proposed models used 1) fewer deep layers, 2) dynamic precision, 3) pruned topology, and 4) layer-shared Intellectual Property (IP).
To further speed up the inference, these entries downsized the images in the CPU before FPGA processing.
In order to compensate the limited on-chip BRAM, these entries utilized the off-chip 512MB DRAM to store the intermediate data between network layers.

\begin{figure}%[htb]
\centering
  \includegraphics[width=0.95\columnwidth]{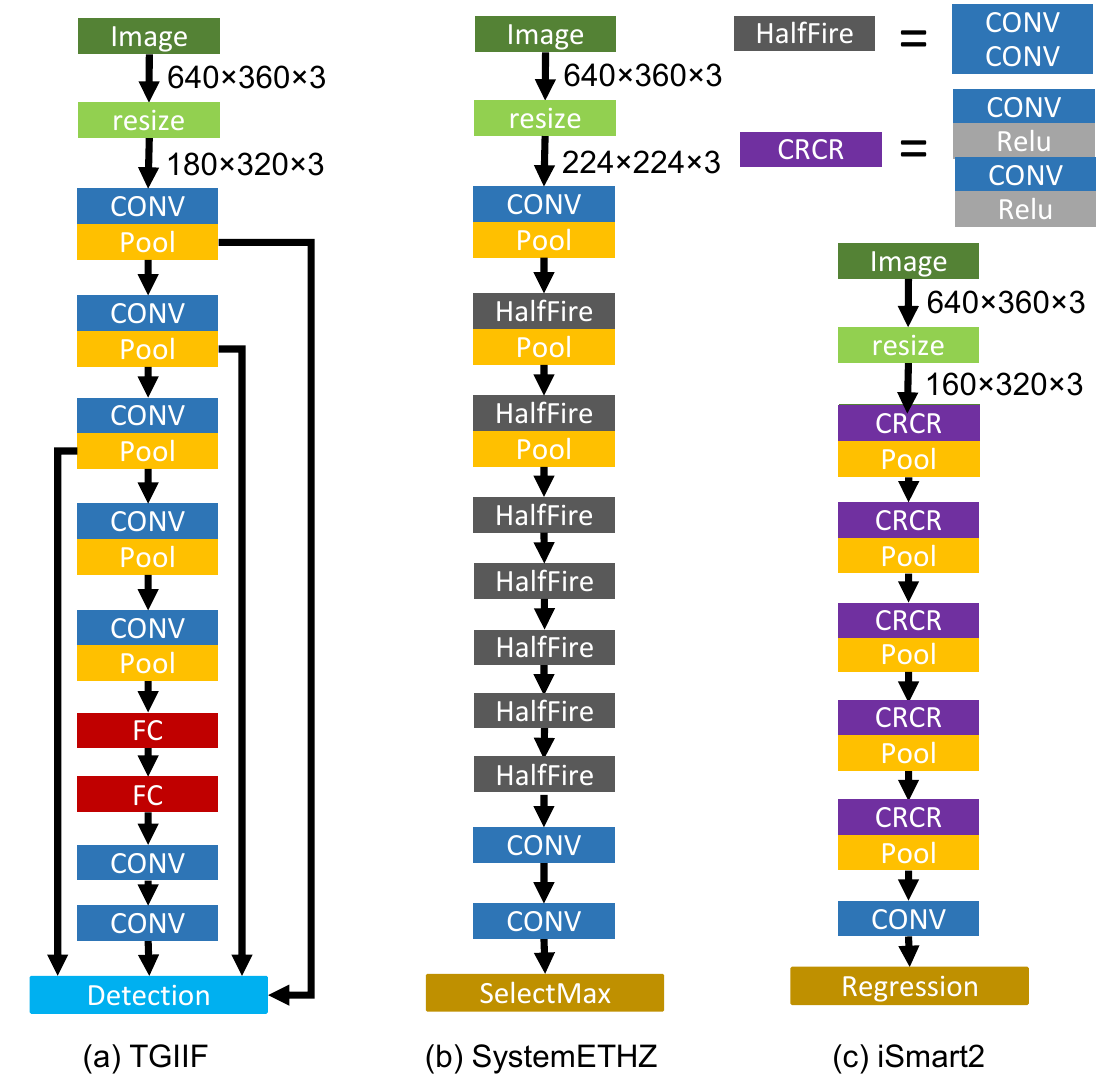}
  \caption{Neural network structure of the top-3 FPGA entries: (a) TGIIF, (b) SystemETHZ, and (c) iSmart2.}
  \label{fig:Network_FPGA}
\end{figure}

The entry \emph{\textbf{TGIIF}} proposed an optimized SSD network.
It downsized the SSD network topology by removing the last two convolutional layers and quantized the network parameters to eight-bit fixed point.
After modifying the network depth, it also pruned and fine-tuned the resultant network, making it with 14.2x less parameters and 10x less operations.
The network topology proposed by \emph{{TGIIF}} is shown in Fig. \ref{fig:Network_FPGA}(a).
% to increase the speed from 1.5FPS to 12FPS suffering only 3.6\% precision loss from 65.7\% to 62.1\%.

%\begin{figure}[htb]
%\centering
  %\includegraphics[width=0.95%\columnwidth]{./figure/TGIIF.pdf}
  %\caption{TGIIF network topology.}
  %\label{TGIIF}
%\end{figure}

%\begin{figure}[htb]
%\centering
  %\includegraphics[width=0.95%\columnwidth]{./figure/SystemsETHZ.pdf}
  %\caption{SystemsETHZ network topology.}
  %\label{SystemsETHZ}
%\end{figure}

%\begin{figure}[htb]
%\centering
  %\includegraphics[width=0.95%\columnwidth]{./figure/iSmart2.pdf}
    %\caption{iSmart2 network topology.}
  %\label{iSmart2}
%\end{figure}

The entry \emph{\textbf{SystemsETHZ}} proposed a variation of SqueezeNet+Yolo networks.
It reduced the fire layer in SqueezeNet to half and binaried the fire layer, and introduced a deep network consisting of 18 convolutional layers where the halfFire layers are binary.
%a total of eighteen binary convolution layers.
The network topology proposed by \emph{{SystemsETHZ}} is shown in Fig. \ref{fig:Network_FPGA}(b).
It also adopted dynamic precision weights among different layers: five-bit fixed point parameters in all activation layers, eight-bit fixed point parameters in the first convolutional layer, and binary weights in all fire layers.
With dynamic precision, it reduced the weight size to 64 KB and the number of multiplication operations to 154 millions.

% achieving a speed of 28.6FPS and precision of 49.19\%.

The entry \emph{\textbf{iSmart2}} proposed a variation on MobileNet+Yolo networks.
It introduced a hardware-friendly network, consisting of multiple depth-wise separable convolution kernels. In each kernel, it adopted a convolution-Relu-Convolution-Relu topology.
The network topology proposed by \emph{{iSmart2}} is shown in the Fig. \ref{fig:Network_FPGA}(c).
By calling one depth-wise separable convolution IP when processing different layers, the proposed network achieved a relatively low resource utilization with minimum look-up-table and Flip-Flop usage.
% It achieved a precision of 57.33\% and speed of 7.35FPS.

The entry \emph{\textbf{traix}} proposed a varied SSD network with sixteen-bit fixed-point parameters.
The entry \emph{\textbf{hwac\_object\_tracker}} proposed a varied Tiny YOLO network topology with half-precision floating point parameters.
The entry \emph{\textbf{Lilou}} proposed a binaried VGG16 network topology with less pooling layers, thus retaining the inference data flow on the FPGA.
The entry \emph{\textbf{Qiu's Team}} proposed a varied PYNQ-BNN \cite{bnnpynq}, adopting 2-bit precision for all layers' parameters.

The FPGA resource utilization of all the entries is shown in Table~\ref{FPGA_resource}.
For entries that adopted parameters with sixteen or eight bit-width (\emph{{TGIIF, SystemsETHZ, iSmart2, traix, hwac\_object\_tracker}}), the DSP utilization is close to 90\%. The top entry \emph{{TGIIF}} achieved a 100\% DSP utilization.
For entries that adopted parameters with one or two bit-width (\emph{{Lilou, Qiu's Team}}), the DSP utilization is as low as 12\% (\emph{{LiLou}}), while the LUT utilization can be as high as 100\% (\emph{{Qiu's Team}}).
%{\color{red}{I may add more?}}

\begin{table}[htbp]
\caption{Resource utilization of FPGA entries}
\label{FPGA_resource}
\centering
\begin{tabular}{ccccc}
\hline
Entries     & LUTs    & Flip-Flop & BRAM    & DSP     \\%\hline
            & (53K)     & (106K)      & (630KB)   & (220)     \\ \hline
TGIIF       & 83.89\% & 54.24\%   & 78.93\% & 100\%   \\
SystemsETHZ & 88\%    & 62\%      & 77\%    & 78\%    \\
iSmart2     & 63\%    & 22\%      & 95\%    & 86\%    \\
traix       & 90\%       & -         & 90\%       & 90\%       \\
hwac\_object\_tracker        & 85.02\% & 41.51\%   & 42.14\% & 87.73\% \\
Lilou       & 75\%    & 38\%      & 98\%    & 12\%    \\
Qiu's Team      & 100\%   & 61\%      & 96\%    & 23\%    \\ \hline
\end{tabular}
\end{table}

% \begin{table}[htbp]
% \caption{Top three entries performance}
% \label{FPGA_performance}
% \centering
% \begin{tabular}{cccc}
% \hline
% Entries     & IoU      & Power(mW) & FPS     \\ \hline
% TGIIF       & 0.623798 & 4200      & 11.9553 \\
% SystemsETHZ & 0.491926 & 2450      & 25.9678 \\
% iSmart2     & 0.5733   & 2590      & 7.3488  \\ \hline
% \end{tabular}
% \end{table}

\section{Results and Analysis}

In this section, we will discuss and analyze the result with respect to method, category, object size, image brightness, and amount of information for both GPU and FPGA entries.
As power and throughput are determined once the method is chosen, we focus on the detection accuracy in this section.

\subsection{Results of GPU Entries}

\subsubsection{Overall Results}

\begin{figure}
\centering
  \includegraphics[width=1\columnwidth]{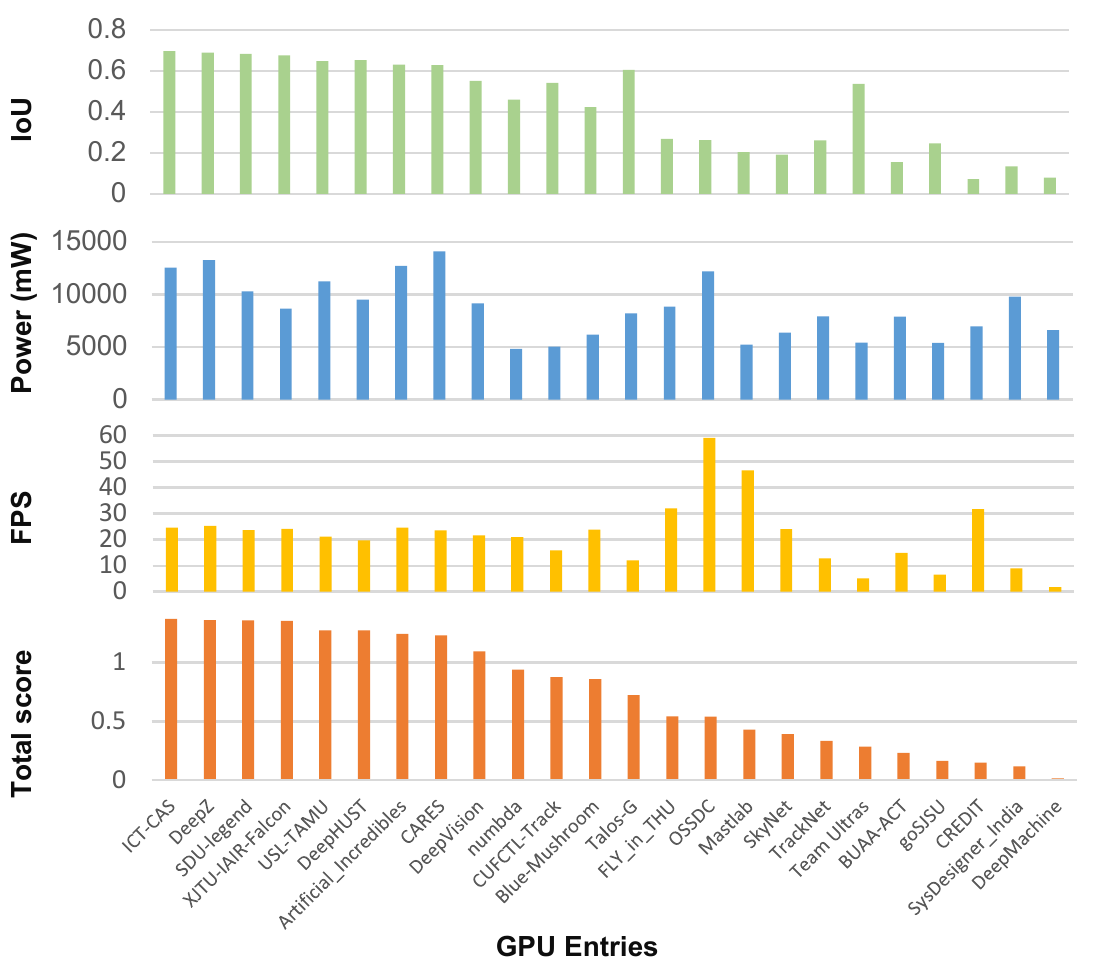}
  \caption{Overall results of GPU entries. Note that the entries are ranked from high to low in the horizontal axis. The details of the scores can be found on the website of the challenge.}
  \label{fig:GPU_method}
\end{figure}

As shown in Fig. \ref{fig:GPU_method}, the optimal IoU, power and FPS are 0.6975, 4834mW, and 58.91 FPS, respectively.
Note that all the top-3 entries (ICT-CAS, DeepZ, SDU-Legend) have high IoUs and FPS.
Only high IoU or FPS is not enough to achieve a high total score/good ranking.
For example, the entry OSSDC obtains a rather high FPS and a low IoU, which only ranks the 15th.
While another entry Talos-G achieved a high IoU and a low FPS (lower than 20), which triggers the penalty as discussed in Equation (\ref{penalty}) and only ranks the 14th.

Actually all the top-8 entries get IoUs higher than 0.60 and throughput higher than 20 FPS.
Moreover, their values are both very close to each other which shows rather fierce competition among the top entries.
Compared with IoUs, the power has a less influence on the final ranking.

\begin{figure}
\centering
  \includegraphics[width=1\columnwidth]{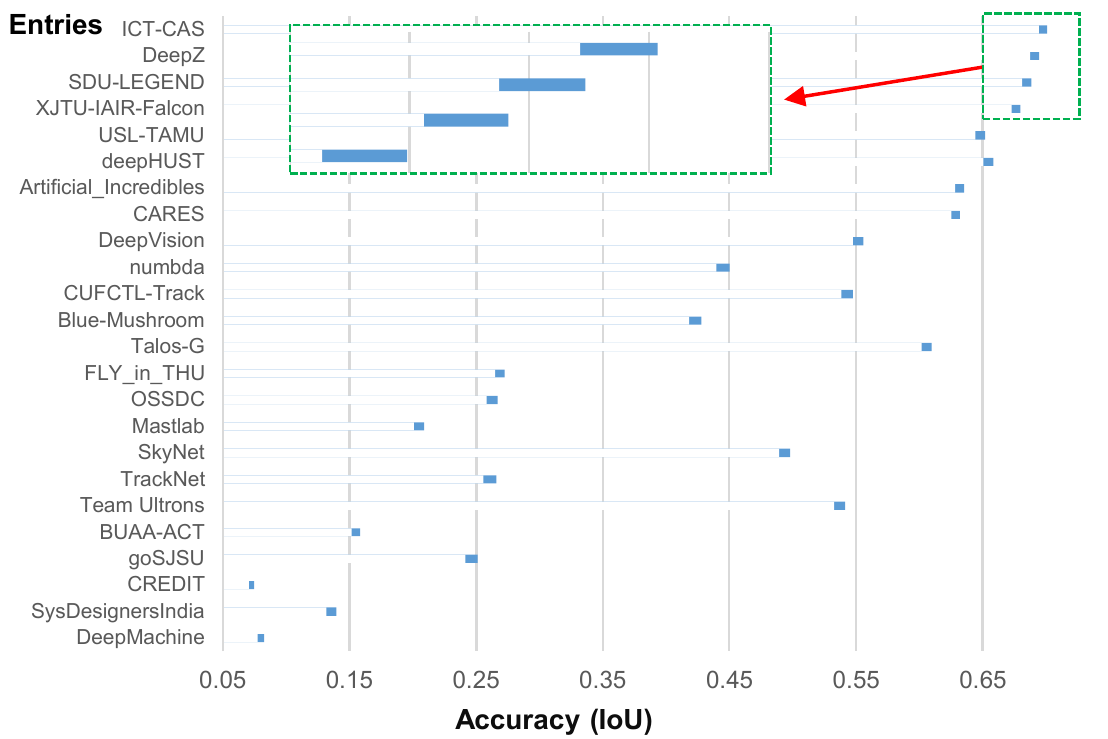}
  \caption{Statistical significance analysis using bootstrapping with 99.9\% confidence intervals for GPU entries. Note that the entries are ranked from high to low in the vertical axis.}
  \label{fig:gpu_statistics}
\end{figure}

In order to analyze whether results of different entries are statistically significantly different from each other, statistical significance analysis is performed.
The bootstrap method is adopted here which is also employed by PASCAL VOC \cite{everingham2010pascal} and ImageNet \cite{russakovsky2015imagenet}.
In each bootstrap round, M images are sampled with replacement from the available M testing images and the average IoU for the sampled images is obtained for one entry.
The above process iterates for each entry until reaching the pre-defined bootstrapping round.
With the results obtained from all the bootstrapping rounds, the lower and upper $a$ fraction are discarded, and the range of the remaining results are the 1-2a confidence interval.
We set the number of bootstrapping rounds to 20,000 and $a$ to 0.0005 (99.9\% confidential interval), and the final results of the entries are shown in Fig. \ref{fig:gpu_statistics}.
It can be observed that almost all the entries are statistically significantly different from each other, and the difference of the top-4 entries are also very obvious even with minor differences.

\subsubsection{Detection Results by Category}\label{gpu_category}

\begin{figure*}
\vspace{-6pt}
  \includegraphics[width=1\textwidth]{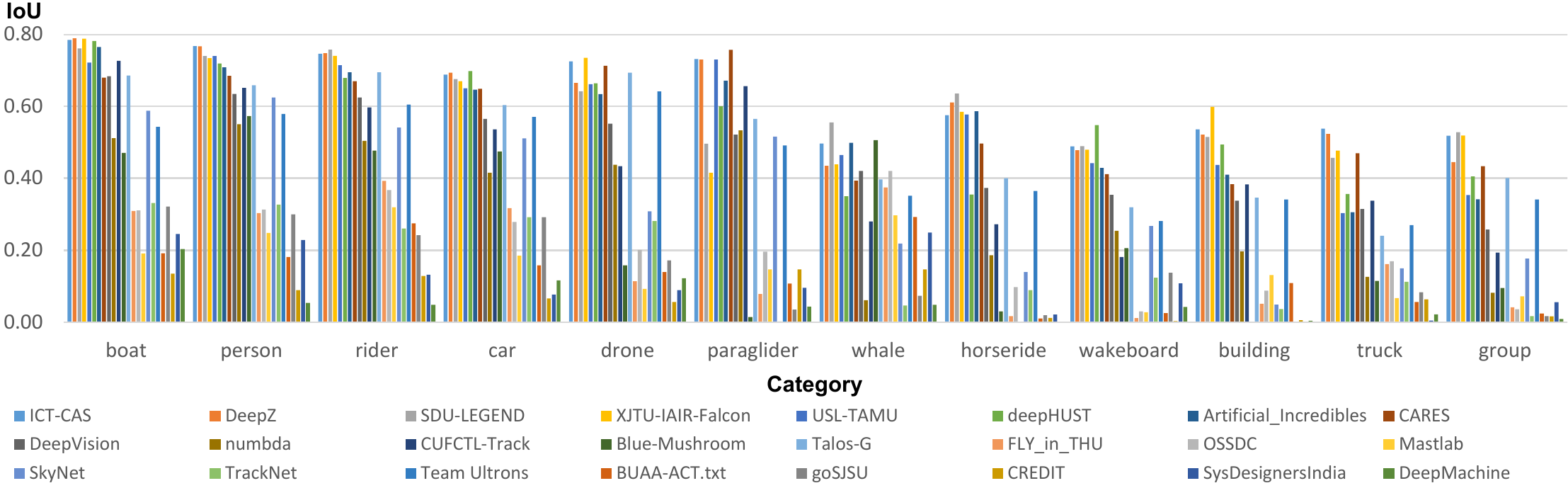}
  \vspace{-20pt}
  \caption{Detection accuracy of GPU entries with respect to image category. Note that the category is sorted in a ranked order and the left-most one is the one with the highest average total score.}
    \vspace{-1pt}
  \label{fig:GPU_class}
  \vspace{-6pt}
\end{figure*}

As shown in Fig. \ref{fig:GPU_class}, the category \textit{boat} is with the highest detection accuracy as it contains moderate quantity of images and its object size is relatively larger than other categories as shown in Fig. \ref{fig:dataset}.
The category \textit{person}, \textit{rider}, and \textit{car} are with a high accuracy as their image quantities are large which can provide a large variety of the object for training as shown in Fig. \ref{fig:dataset_category}.
The category \textit{drone} and \textit{paraglider} also get high accuracy (though their image quantities are small) which is due to the fact that their backgrounds are usually simple such as the sky and their structures are also very special compared with others.

The rest six categories are with relatively lower accuracy as their image quantities are small as shown in Fig. \ref{fig:dataset_category}.
Furthermore, the category \textit{whale} has a very low contrast between the object and the background (the sea) as shown in Fig. \ref{fig:dataset}.
The category \textit{building} is rather challenge as there exists many similar objects which results in a very low accuracy.
The category \textit{group} gets the lowest accuracy as there also exists multiple similar objects which makes it very hard to detect the right one.

\begin{figure*}
\vspace{-6pt}
  \includegraphics[width=1\textwidth]{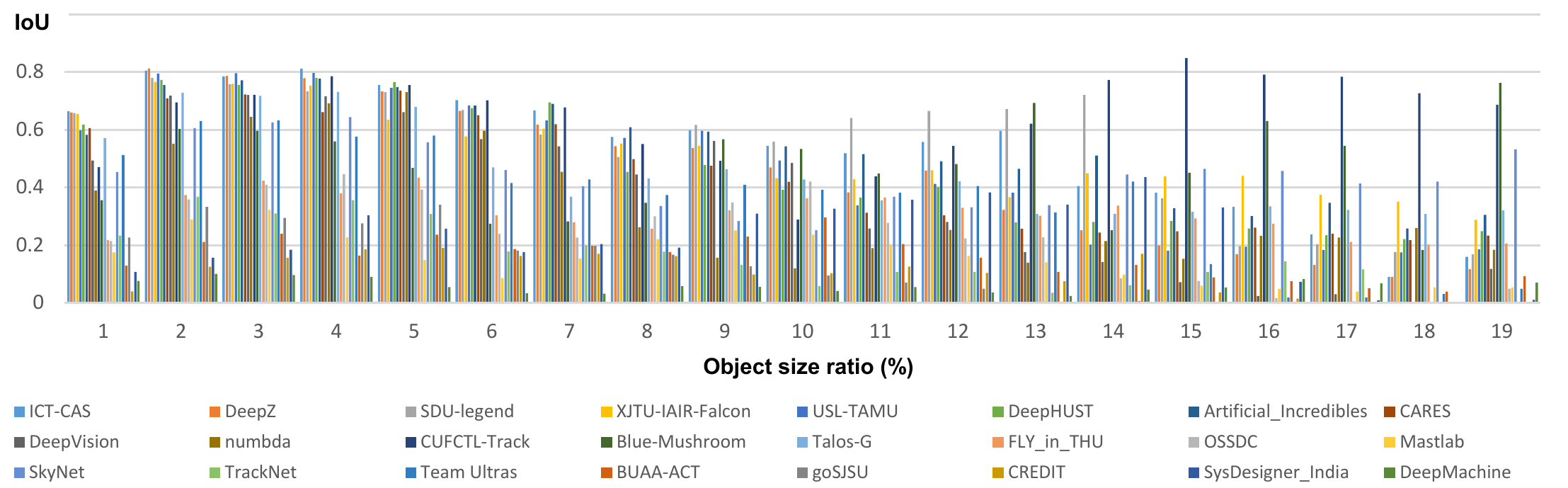}
  \vspace{-20pt}
  \caption{Detection accuracy of GPU entries with respect to size.}
    \vspace{-1pt}
  \label{fig:gpu_size}
  \vspace{-10pt}
\end{figure*}

\begin{figure*}
  \includegraphics[width=1\textwidth]{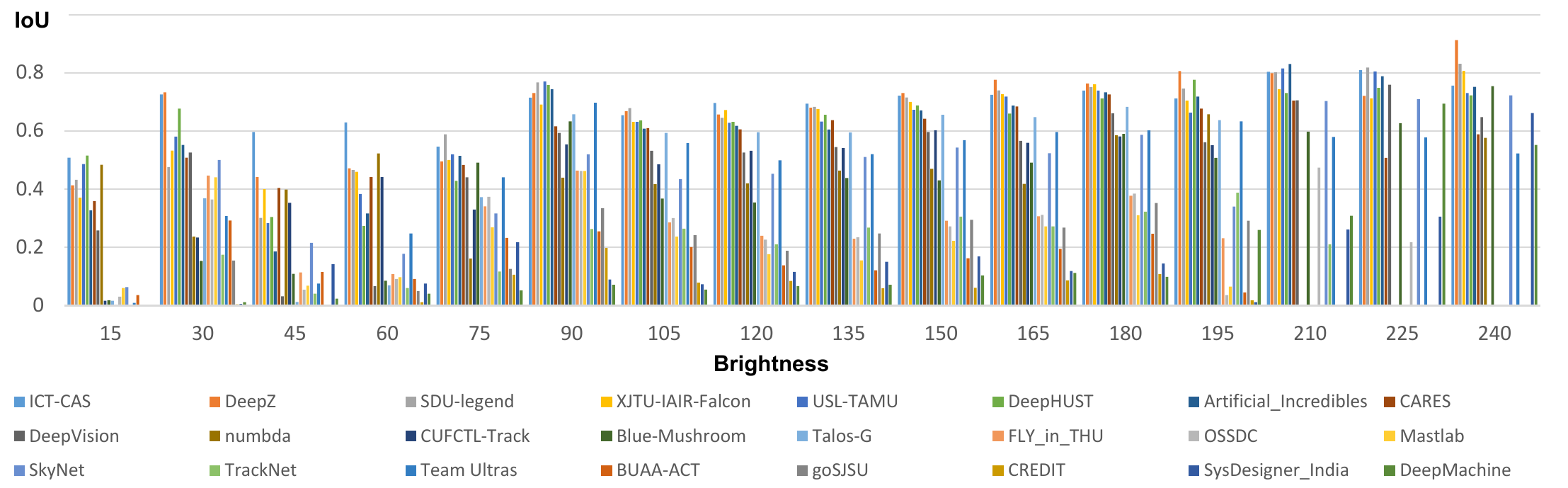}
  \vspace{-20pt}
  \caption{Detection accuracy of GPU entries with respect to brightness.}
    \vspace{-1pt}
  \label{fig:gpu_brightness}
  \vspace{-16pt}
\end{figure*}

\begin{figure*}
  \includegraphics[width=1\textwidth]{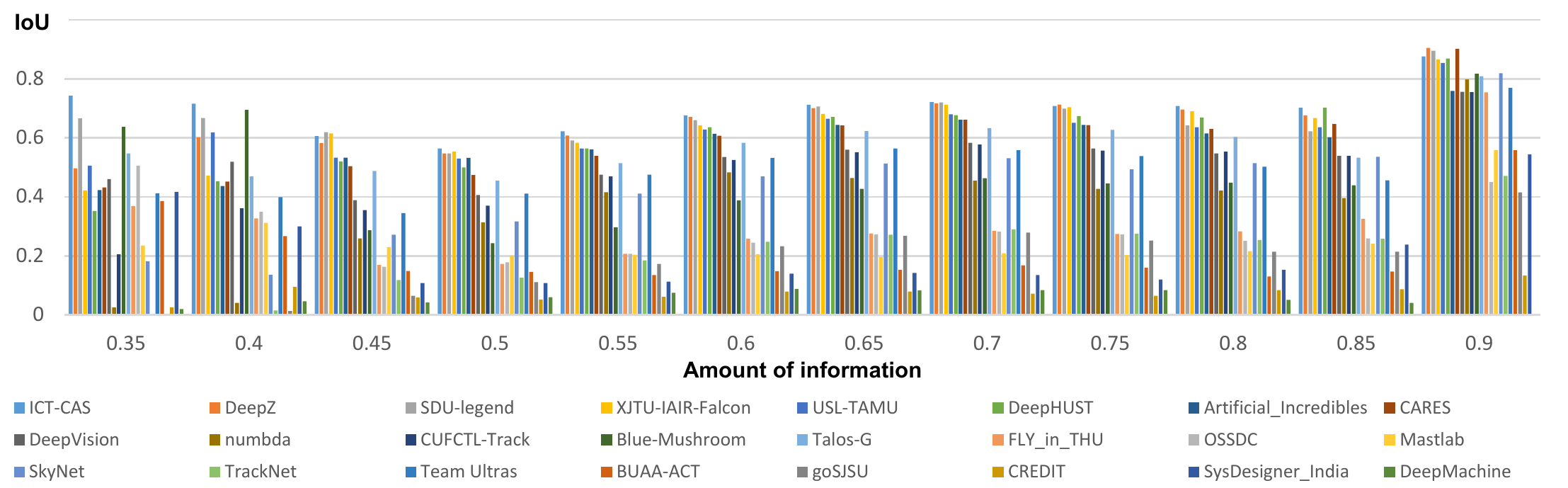}
  \vspace{-20pt}
  \caption{Detection accuracy of GPU entries with respect to amount of information.}
    \vspace{-1pt}
  \label{fig:gpu_texture}
  \vspace{-16pt}
\end{figure*}

\subsubsection{Detection Results by Object Size}\label{gpu_size}

In Fig. \ref{fig:dataset_size} the quantity of images with 1\% object size ratio is larger than that of images with 2\% object size ratio.
However, as shown in Fig. \ref{fig:gpu_size}, the accuracy of images with 1\% object size ratio is much smaller than that of images with 2\% object size ratio.
The main reason is that too small object is much harder to be detected with high accuracy.
The accuracy of the images with 2\%-5\% object size ratios is almost the same and relatively higher than others as their corresponding image quantity is large and the object size is moderate.
However, when the object size ratio increases from 6\%, the accuracy decreases as the corresponding training image quantity is much lower than others.

\subsubsection{Detection Results by Image Brightness}\label{gpu_brightness}
As shown in Fig. \ref{fig:gpu_brightness}, there is a almost linear trend between the accuracy and the image brightness.
As the image brightness increases, the accuracy also increases.
However, if the accuracy is directly correlated with the image brightness, the accuracy with a brightness of about 135 should be the highest as the image quantity with this brightness is the largest.
Thus,
%there is no directly correlation between the image brightness and the accuracy.
we tend to believe that higher brightness will make the object more clear to show its features which will make it relatively easier to be detected with high accuracy.

\subsubsection{Detection Results by Amount of Information}\label{gpu_texture}

As shown in Fig. \ref{fig:gpu_texture}, images with larger amount of information tend to have higher accuracy as larger amount of information contains more features which are relatively easier to be detected.
The images with amount of information of 0.35-0.40 get a slightly higher accuracy than that with amount of information of 0.45-0.50 which may caused by that these images are with unique characteristics and a small quantity.
We can also notice that the entry ICT-CAS can get a much higher accuracy for images with amount of information of 0.35-0.40 than others, and get relatively lower accuracy for images with amount of information of 0.90 than other top-3 entries.

\begin{figure}
\centering
  \includegraphics[width=1\columnwidth]{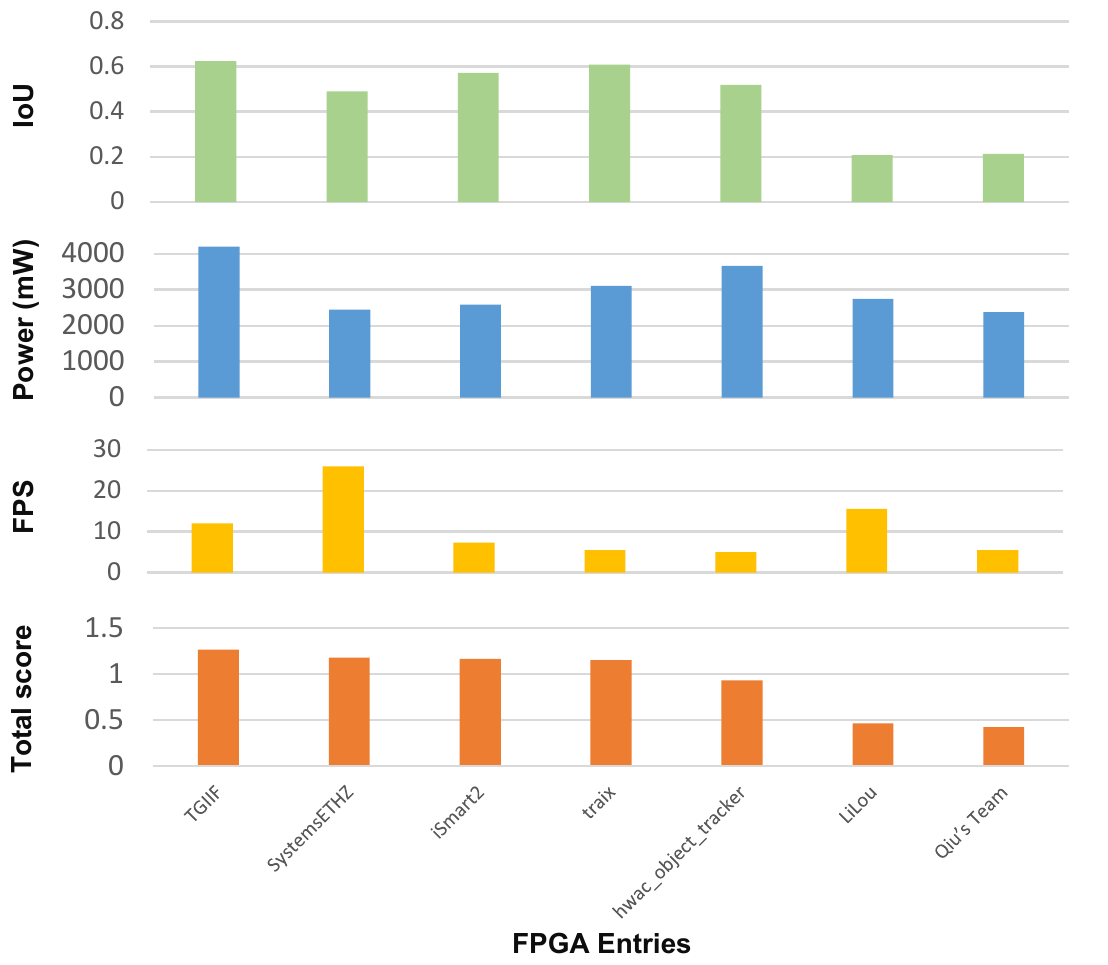}
    \caption{Overall results of FPGA entries. The details of the scores can be found on the website of the challenge.}
  \label{fig:FPGA_by_method}
\end{figure}

\subsection{Results of FPGA entries}
\subsubsection{Overall Results}

% This section reports and discusses the results of the FPGA task.
% The seven entries were evaluated.
% All entries tackled all of the twelve classes.

Fig.~\ref{fig:FPGA_by_method} summarizes the performance of all FPGA entries.
%The seven entries were evaluated.
Entry TGIIF stands out as the most successful method, obtaining an IoU score of 0.6238 and a throughput of 11.96 FPS. The adopted deep network leads to a high IoU score, and the network pruning ensures the inference throughput.
Entry SystemsETHZ comes second while it achieves a throughput of 25.97 FPS and an IoU score of 0.4919.
Its dynamic precision architecture and binarized layers accelerate the inference throughput at the expense of IoU loss.
iSmart2 comes third with a throughput of 7.35 FPS and an IoU score of 0.5733.
Its uniform-layer deep network design gives rise to IP reuse and fine-grained memory allocation, keeping a good balance between throughput and accuracy.

% The FPGA performance in this challenge is not only influenced by the network topology and depth, but also by the design deployment. The network typologies adopted determine the accuracy of the design and the FPGA design deployment impact the processing speed.

Many entries achieved IoUs of around 0.5-0.6, such as entry $traix$ which achieves an IoU of 0.6100 with sixteen-bit fixed-point network parameters.
Despite the success in accuracy, there is a quite range in inference throughput.
As mentioned in section 2.2 and 4.2, the FPGA platform has 630KB on-chip BRAM, which is not sufficient to retain all parameters and  output feature maps of layers.
%While achieving a high accuracy, the limited on-chip BRAM may not be able to retain all parameters and layer output feature maps on-chip.
Thus, a frequent data transfer between DRAM and FPGA are observed in all designs, and then keeping a proper balance between accuracy and network size is very important in this challenge.
%Thus, keeping a balance between accuracy and network size is very important in this challenge.
%Base on some succeeded networks, t
The standout top-3 entries adopted network pruning, parameters binarization and IP-reuse to ensure the inference throughput.

\begin{figure}
\centering
  \includegraphics[width=1\columnwidth]{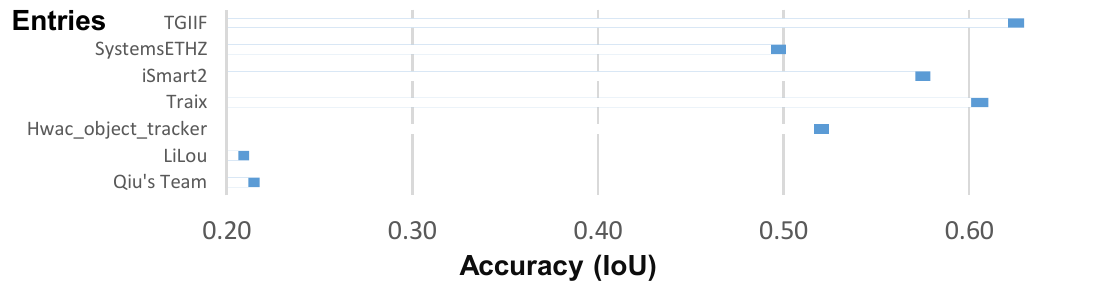}
  \caption{Statistical significance analysis using bootstrapping with 99.9 confidence intervals for FPGA entries.}
  \label{fig:FPGA_statistics}
\end{figure}

The statistical significance analysis is shown in Fig. \ref{fig:FPGA_statistics}.
It can be easily noticed that all the entries are statistically significantly different from each other, and the difference of the top-3 entries are even larger and there is also moderate difference between their accuracies.

\subsubsection{Detection Results by Category}\label{fpga_category}
\begin{figure}%[htb]
\centering
  \includegraphics[width=1\columnwidth]{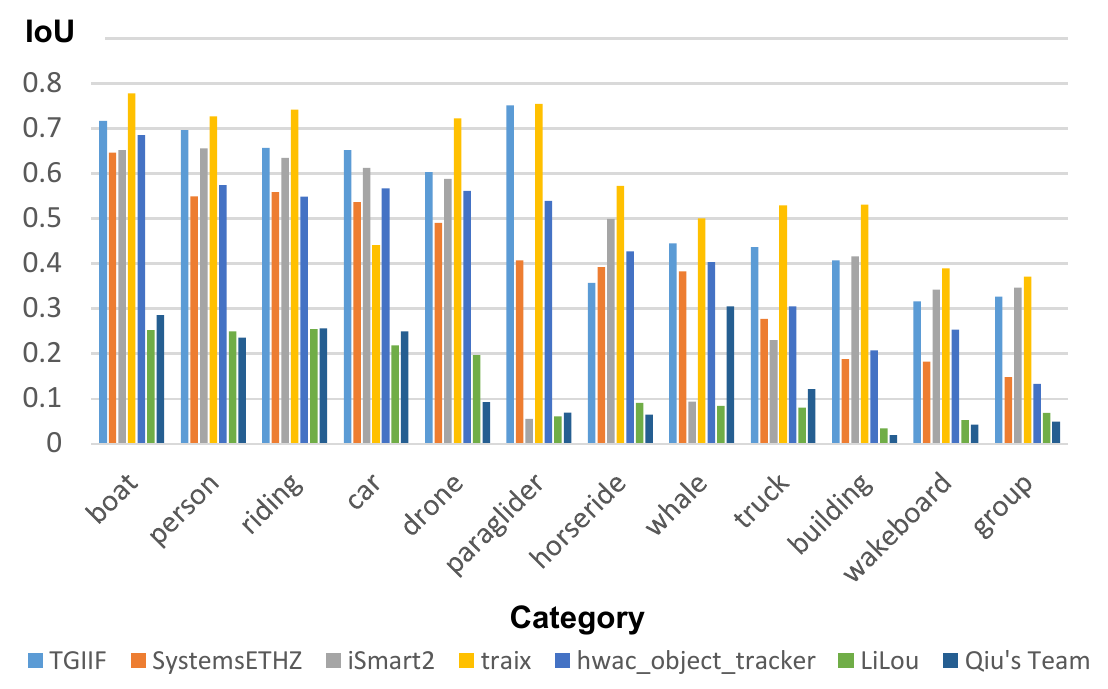}
    \caption{Detection accuracy of FPGA entries with respect to image categories. Note that the category is sorted in a ranked order and the left-most one is with the highest average total score.}
  \label{fig:FPGA_by_class}
\end{figure}

Fig.~\ref{fig:FPGA_by_class} summaries the results obtained according to object category. The categories are listed in sequence of average IoU.
There is drastic variation among categories and in different entries, in which, $boat$ keeps the most promising average IoU of 0.5734 and $group$ with an average IoU of 0.2060 is the most challenging category to be detected.
On average IoU, the most promising categories are $boat$, $person$, $rider$, and $car$, while those four categories have moderate quantities of images and their object sizes are relatively larger than other categories.
Among those four categories, $boat$ stands out as the most promising category while it contains 5k (the least) training images compared with $person$ which contains 28k (the most) training images.
This is due to the fact that the background in the $boat$ category is simple (the sea) and the target is distinct, while the $person$ category contains numerous noise such as trees, grass, and even other non-objected persons.
The $rider$ stands ahead of $car$ while the $car$ category contains 25k training images and $rider$ contains only 1.6k training images.
This is due to the fact that the $rider$ category is captured from a relatively close view while $car$ category is captured from a distant view.

For these challenging categories such as $whale$, $truck$, $building$, and $wakeboard$, there is an approximately 20\% accuracy gap between these categories and the promising categories for both maximum IoU and average IoU.
%Rather than the low image quantity, the object features may also worsen the detection accuracy.
Not only the low image quantity, but also the complex object features worsen the detection accuracy.
The features of these four challenging categories can be summarized as follows: the category $whale$ contains blurry object and background, while categories $truck$, $wakeboard$, and $building$ have tiny objects or similar objects to the target objects such as the same color and shadow.

\subsubsection{Detection Results by Object Size}\label{fpga_size}
\begin{figure}%[htb]
\centering
  \includegraphics[width=1\columnwidth]{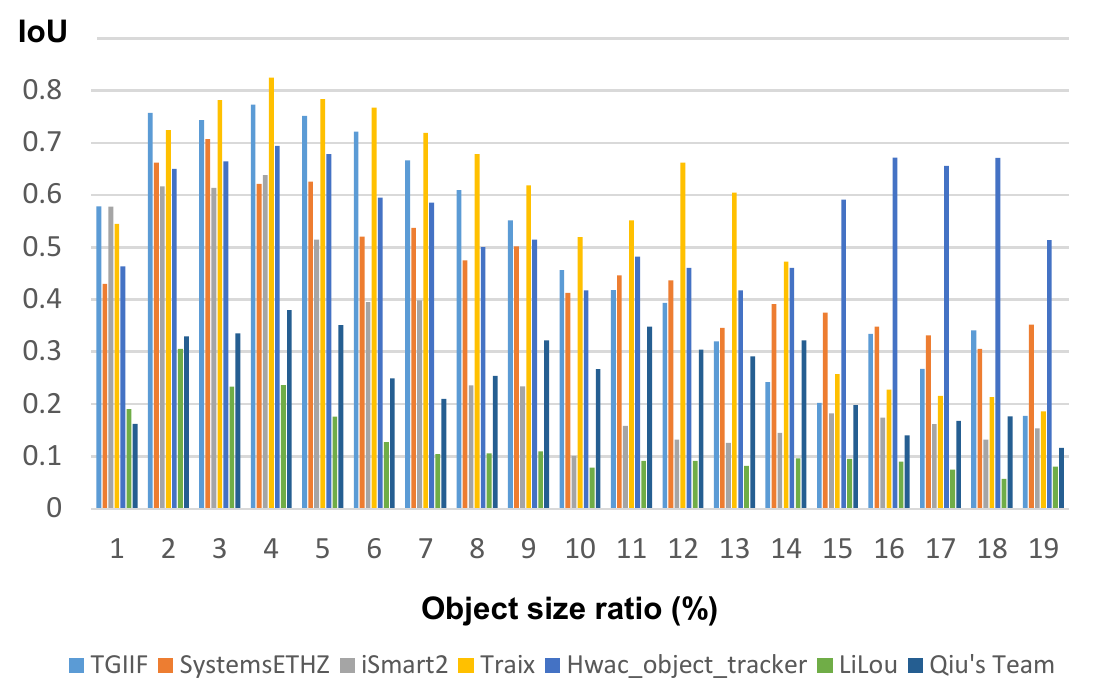}
  \caption{Detection accuracy of FPGA entries with respect to size.}
  \label{fig:FPGA_size}
\end{figure}

% Images taken from overlook view have features such as smaller size, different point of view and more irrelevant objects.
Fig. \ref{fig:FPGA_size} summaries the overall IoU of the seven entries with the target size.
%The sequence of the target size is sequenced as it is shown in Fig.\ref{fig:dataset_size}.
It can be observed than the most promising target size is located at 2\%\~{}6\%, and 4\% stands out as the most promising target size.
While the training images with target size lower than 2\% count the largest quantity of the dataset, its accuracy is inferior compared with images with a target sizen of 2\%\~{}6\%.
The accuracy of images with less than 2\% target size is slightly superior than those images with 6\%\~{}10\% target size, while the number of images with less than 2\% target size is almost 100x more than those with 6\%\~{}10\% target size.
The overall accuracy starts downgrading after ratio 4\% while entry $hwac\_object\_tracker$ keeps a relatively high accuracy.
This may due to the fact that its framework has a good capability in extracting features of big objects.


\subsubsection{Detection Results by Image Brightness}\label{fpga_brightness}
\begin{figure}%[htb]
\centering
  \includegraphics[width=1\columnwidth]{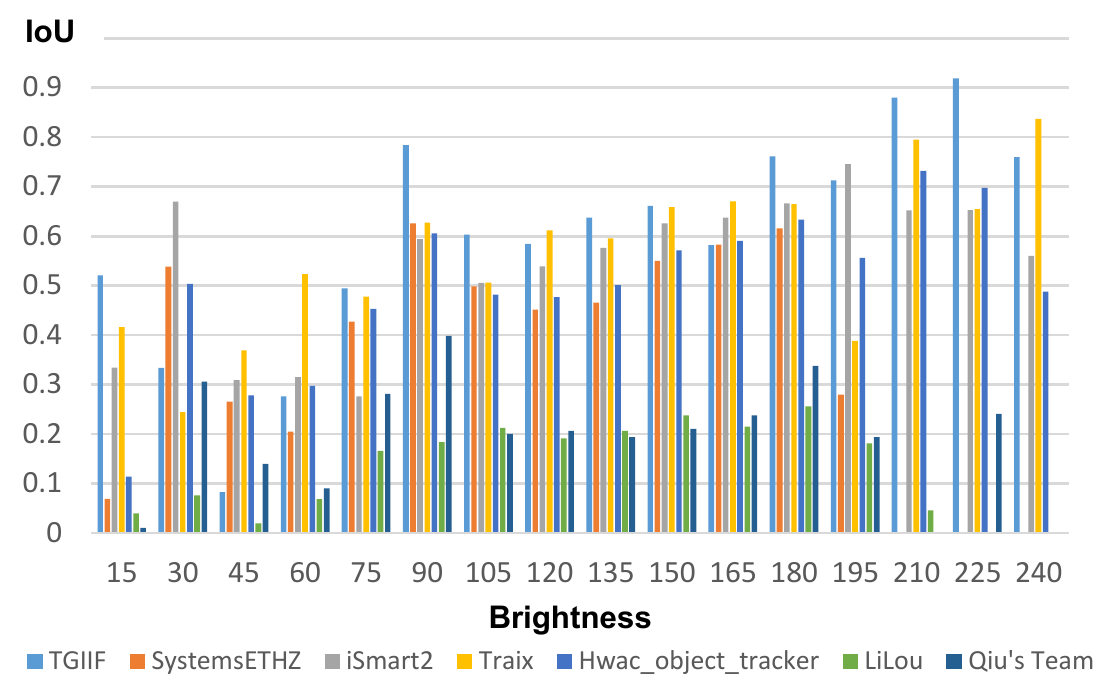}
  \caption{Detection accuracy of FPGA entries with respect to brightness.}
  \label{fig:FPGA_brightness}
\end{figure}

%The distribution of image brightness is similar to a Gaussian distribution as shown in Fig.~\ref{fig:dataset_brightness}.
In Fig.~\ref{fig:FPGA_brightness}, most of the training images are captured with the brightness between 105 to 165.
%The accuracy of images with different brightness is summarized in Fig.~\ref{fig:FPGA_brightness}.
However, the overall accuracy of seven entries doesn't follow the same trend as shown in Fig. \ref{fig:FPGA_brightness}.
The optimum point first arises when the brightness reaches 90 from 30.
And the trend of accuracy starts slowly climbing until 195.
After 195, the detection results show a drastic variation among entries due to the small quantities of the corresponding training images.
Thereby, it can be deducted that the detection accuracy is correlated to brightness which is the same as that for GPU entries.

\subsubsection{Detection Results by Amount of Information}\label{fpga_texture}
\begin{figure}%[htb]
\centering
  \includegraphics[width=1\columnwidth]{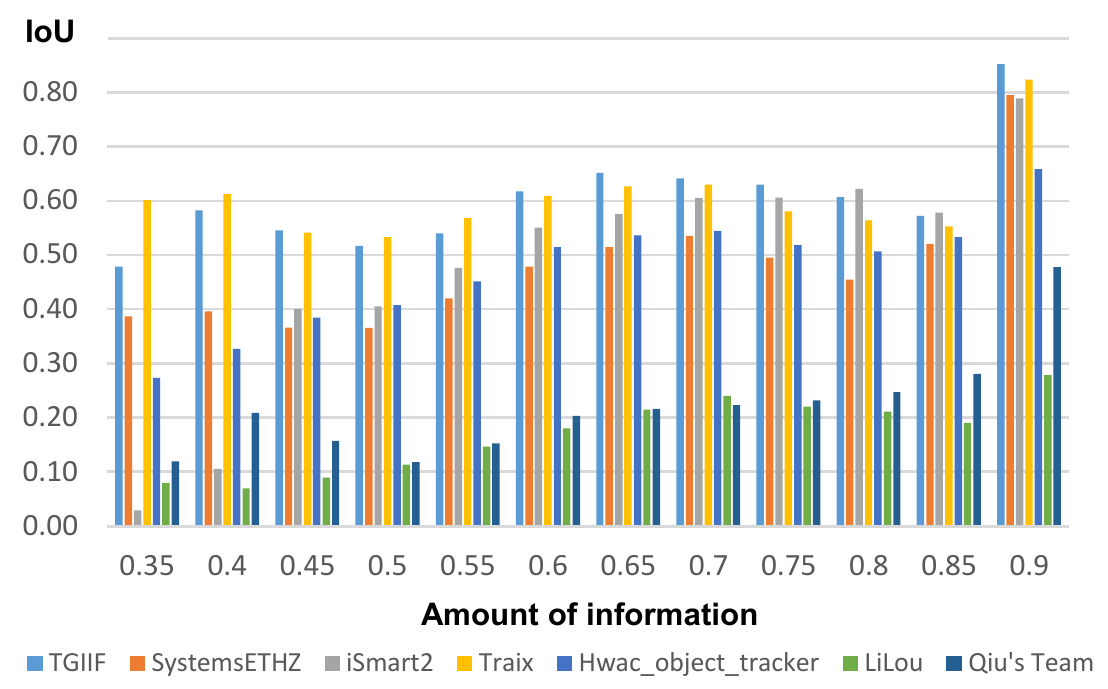}
  \caption{Detection accuracy of FPGA entries with respect to amount of information.}
  \label{fig:FPGA_texture}
\end{figure}

Fig.~\ref{fig:FPGA_texture} summaries the accuracy of all entries with the amount of information.
It can be observed that the overall accuracy trends to increase until the amount of information reaches 0.65, which is the same with the trend shown in Fig.~\ref{fig:dataset_texture}.
However, for images with an amount of information of 0.9, their accuracy is the highest though their quantities are the smallest.
Thus, it can be deducted that larger amount of information tends to achieve higher accuracy which is the same as that for GPU entries.

%\begin{table}[htbp]
%\caption{Performance of different methods}
%\label{FPGA_performance}
%\centering
%\begin{tabular}{ccccc}
%\hline
%Entries      & IoU      & Power(mW) & Energy(mWh) & FPS     \\ \hline
%TGIIF        & 0.623798 & 4200      & 5087        & 11.9553 \\
%SystemsETHZ  & 0.491926 & 2450      & 1432        & 25.9678 \\
%iSmart2      & 0.5733   & 2590      & 5124        & 7.3488  \\
%traix        & 0.610656 & 3110      & 8406        & 5.4448  \\
%hwac         & 0.520074 & 3660      & 10767       & 4.9354  \\
%Lilou        & 0.2084   & 2750      & 2546        & 15.6059 \\
%Qiu's entry   & 0.214842 & 2380      & 6404        & 5.4547  \\
%ApproxiTrack & 0.33766  & 2620      & 61200       & 1.0404  \\ \hline
%\end{tabular}
%\end{table}

%\begin{figure}[htb]
%\centering
%  \includegraphics[width=1\columnwidth]{./figure/FPGA_by_class_max.pdf}
%    \caption{Detection accuracy of GPU FPGA in category--to be deleted.}
%  \label{fig:FPGA_by_class_max}
%\end{figure}

\begin{figure*}
\vspace{-6pt}
  \includegraphics[width=1\textwidth]{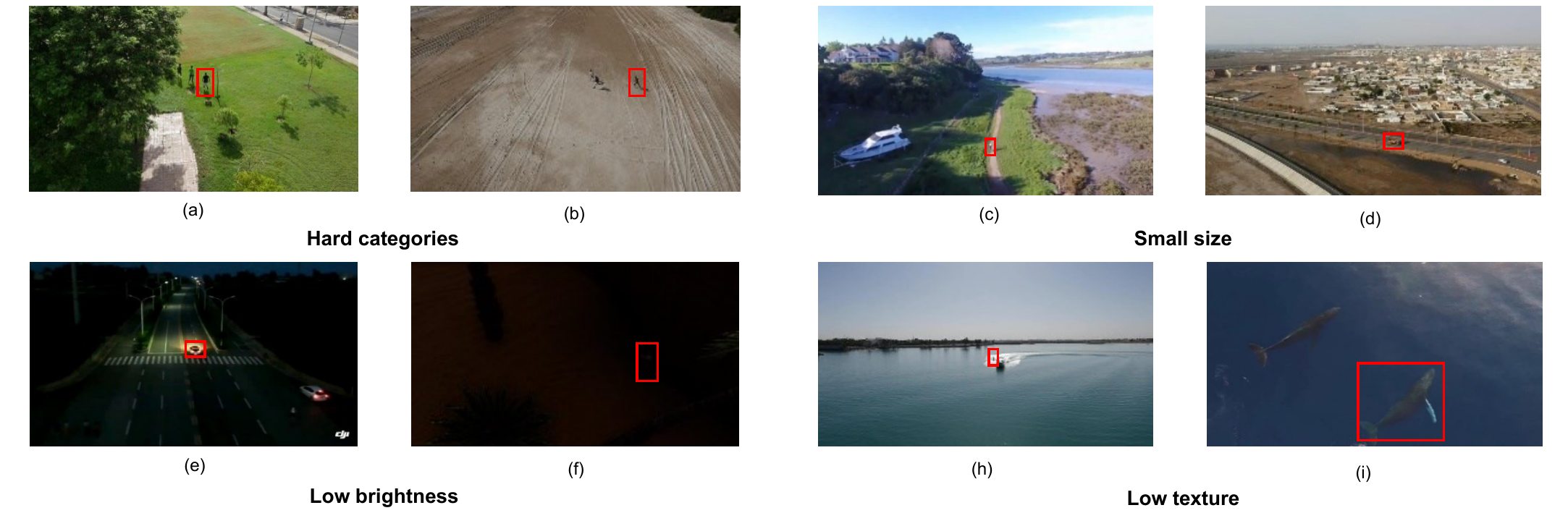}
  \vspace{-20pt}
  \caption{Hard examples with hard categories, small size, low brightness, and low amount of information in the challenge.}
    \vspace{-1pt}
  \label{fig:hardExample}
  \vspace{-10pt}
\end{figure*}
\subsection{Hard Examples and Lessons}

The results for GPU and FPGA entries show almost the same phenomena and trends as they adopted the same method (deep neural networks), and their hard-to-detect images are also almost the same.
Fig. \ref{fig:hardExample} shows some hard examples with hard categories, small sizes, low brightness, and low amount of information for both GPU and FPGA entries.
The most challenge category is \textit{group} as discussed in Section \ref{gpu_category} and Section \ref{fpga_category}.
As shown in Fig. \ref{fig:hardExample}(a) and (b), the objects in the two images are very small, and there are several similar objects around, which makes accurate detection rather hard.
The two images (c) and (d) contain very small objects (\textit{rider} in image (c) and \textit{truck} in image (d)), both of which are hard to be recognized even by humans.
Images (e) and (f) are two images with very low brightness, and the objects are very blurry without clear outlines.
With the context in image (e), humans can infer that the object is a car.
However, the object in image (f) is very hard for humans to recognize as the light is too dim.
The objects in image (h) and (i) are with very low average amount of information, which is also hard to recognize.
Image (h) contains a very small object which is almost invisible and has very limited information, while image (i) has a object which is large however with smooth surfaces and unclear boundaries with the sea.
Within all the hard examples, it can be observed that almost all the images are within very small objects.
In fact small objects are common for UAV applications which is the major challenge for accurate object detection.
Furthermore, similar objects (\textit{persons}, \textit{rider}, \textit{buildings}, \textit{boats}, \textit{cars}, etc.) and special scenarios (\textit{wake board}, operation at night) add more difficulties.

With the above hard examples and previous results discussed, we present the learned lessons as follows.
First, FPGA is much more energy efficient than GPU.
Though FPGA achieves a relatively lower FPS than GPU,  it can achieve almost the same accuracy but with only 1/3-1/2 energy consumption as that of GPU, which is promising for long-term UAV applications.
Second, object detection from UAV views in real world is complicated.
In the contest, there are many images that can not be accurately detected by all the entries.
Dividing the task into well-defined sub-tasks for specific scenarios may improve the performance.
Third, more data is preferred for more accurate detection.
In the challenge we find that many objects get a low detection accuracy when their brightness or view-angle changes.
Training images with more diversity (scale, view-angel, etc.) of the object will further improve the overall accuracy.

\section{Conclusion}
In this paper we present the DAC-SDC low power object detection challenge for UAV applications.
Specifically, we describe the unique dataset from UAV views, and give a detailed discussion and analysis of the participating entries and their adopted methods.
We also discuss the details of methods proposed by the LPODC entries for efficient processing for UAV applications.
The result analysis provides a good practical lesson for researchers and engineers to further improve energy-efficient object detection in UAV applications.

% or


%\appendices

%\section{}

\ifCLASSOPTIONcompsoc
  \section*{Acknowledgments}
\else
  \section*{Acknowledgment}
\fi

The authors would like to thank DJI for providing the dataset, Xilinx and Nvidia for providing free GPU and FPGA platforms, and DAC organizers for their support to the DAC SDC challenge.

\ifCLASSOPTIONcaptionsoff
  \newpage
\fi

\end{document}